\documentclass{article}

\usepackage{amsmath, amsthm, amssymb, amsfonts}
\usepackage{thmtools}
\usepackage{graphicx}
\usepackage{setspace}
\usepackage[margin = 2cm]{geometry}
\usepackage{float}
\usepackage{hyperref}
\usepackage[utf8]{inputenc}
\usepackage[english]{babel}
\usepackage{framed}
\usepackage[dvipsnames]{xcolor}
\usepackage{tcolorbox}
\usepackage{dsfont}
\usepackage{biblatex}
\usepackage{csquotes}
\usepackage{yhmath}
\usepackage{mathtools}
\usepackage{blindtext}
\usepackage{titlesec}
\usepackage{multirow}
\usepackage{longtable}

\colorlet{LightRed}{Red!15}
\colorlet{LightBlue}{Blue!15}
\colorlet{LightGreen}{Green!15}

\declaretheoremstyle[name=Theorem,]{thmsty}

\tcolorboxenvironment{theorem}{colback=LightRed}

\declaretheoremstyle[name=Proposition,]{prosty}

\tcolorboxenvironment{proposition}{colback=LightBlue}

\declaretheoremstyle[name=Definition]{defsty}

\tcolorboxenvironment{definition}{colback = LightGreen}
\usepackage{authblk}

\title{Heating Up Quasi-Monte Carlo Graph Random Features: \\A Diffusion Kernel Perspective}
\author{%
  Brooke Feinberg\thanks{Scripps College, \texttt{bfeinber3229@scrippscollege.edu}}%
  \hspace{0.5cm} 
  Aiwen Li\thanks{University of Pennsylvania, \texttt{aiwenli@wharton.upenn.edu}}%
}
\date{\vspace{0.01cm}%
  {Project Advisor:} Dr. Fred Hickernell\thanks{Illinois Institute of Technology, \texttt{hickernell@iit.edu} }\\[2ex]
  May - August 2024
}

\begin{document}
\maketitle
    \begin{abstract}
        We build upon a recently introduced class of \textit{quasi-graph random features} (q-GRFs), which have demonstrated the ability to yield lower variance estimators of the 2-regularized Laplacian kernel (Choromanski 2023). Our research aims to investigate whether similar results can be achieved with alternative kernel functions, specifically the Diffusion (or Heat), Matérn, and Inverse Cosine kernels. We find that the Diffusion kernel performs most similarly to the 2-regularized Laplacian, and we further investigate graph types that benefit from the previously established \textit{antithetic termination} procedure. In particular, we explore Erdős-Rényi and Barabási-Albert random graph models, Binary Trees, and Ladder graphs with the goal of identifying combinations of a specific kernel and graph type that benefit from antithetic termination. We assert that q-GRFs achieve lower variance estimators of the Diffusion (or Heat) kernel on Ladder graphs. However, the number of rungs on the Ladder graphs impacts the algorithm's performance--further theoretical results supporting our experimentation are forthcoming. This work builds upon some of the earliest Quasi-Monte Carlo methods for kernels defined on combinatorial objects, paving the way for kernel-based learning algorithms and future real-world applications in various domains.
    \end{abstract}
\section{Introduction}
Combinatorial objects refer to structured data that can be represented in terms of discrete objects like graphs, sets, or trees. These objects are more complex than standard numerical data because they encapsulate structural information about the relationships and interactions between various components. By employing a \textit{kernel trick}--a powerful technique for modeling nonlinear relationships using linear learning algorithms--kernels defined on discrete input spaces enjoy a variety of applications, particularly in the realms of bioinformatics, community detection, and more recently, generative modeling \cite{reid_general_2024}.\\\\
Graphs, in particular, serve as one of the most general representations of discrete metric spaces. To measure the “similarity” between nodes, one can employ a kernel function, or \textit{graph kernel}, to implicitly map these combinatorial objects to a higher dimensional feature space. With this technique, the object of key importance is the \textit{Gram Matrix} \(\textbf{K}\in \mathbb{R}^{NxN}\) whose entries enumerate the pairwise kernel evaluations \(\textbf{K} :=[K(x_i, x_j)]^N_{i,j=1}\) \cite{reid_general_2024}. Despite the inherent rigor and success of these kernel-based learning algorithms, the computational time to construct and invert the kernel matrix leads to a notoriously poor \(\mathcal{O}(N^3)\) time-complexity, sparking research into \textit{random features}: a Monte Carlo approach for efficiently approximating \(\textbf{K}\) \cite{reid_general_2024}. \\\\
A recently viable \textit{graph random features} mechanism proposed by Choromanski (2023) generalizes the random features technique to graphs. A key limitation with this approach is that it only addresses a niche class of graph kernels. To remedy this issue, researchers propose a \textit{general graph random features} (g-GRFs) algorithm by introducing a \textit{modulation function}. This approach generalizes the algorithm to an arbitrary class of functions of a weighted adjacency matrix, allowing for a more efficient and unbiased estimator of a much larger class of graph kernels \cite{reid_general_2024}. Furthermore, Choromanski, Reid, and Weller improve the efficiency of g-GRFs by introducing correlated ensembles, or \textit{antithetic walkers}. This technique, coined \textit{quasi-Monte Carlo graph random features} (q-GRFs), yields lower variance estimators of the 2-regularized Laplacian kernel under mild conditions \cite{reid_quasi-monte_2023}. \\\\
In our research, we aim to build upon the foundational work of Choromanski, Reid, and Weller to identify an alternative kernel function or graph type that achieves a similarly low variance estimator for the q-GRFs procedure. Specifically, we compare the previously proposed 2-regularized Laplacian kernel with the Diffusion (or Heat), Matérn, and Inverse Cosine kernels. Additionally, we explore various graph types that benefit from antithetic termination, particularly the Erdős-Rényi and Barabási-Albert random graph models, Binary Trees, and Ladder graphs.\\\\
We assert that q-GRFs yield lower variance estimators of the Diffusion (or Heat) kernel on ladder graphs with $9$ and $10$ rungs, though further theoretical results investigating and supporting this phenomenon are forthcoming. Our work with kernel functions holds practical significance, as identifying a lower variance estimator typically leads to more accurate and reliable results, enhancing the overall performance of kernel-based learning algorithms. This work builds upon some of the earliest quasi-Monte Carlo methods for kernels defined on combinatorial objects, paving the way for future research and applications. 

\section{Background, Motivation, and Related Work}
\subsection{Graph Kernels}
Kernel methods provide powerful techniques for modeling nonlinear relationships using linear learning algorithms \cite{reid_general_2024}. Kernel-based algorithms capture the structure of an input space $X$ via the kernel $K : X \times X \rightarrow \mathbb{R}$ \cite{goos_kernels_2003}. As long as one can identify an appropriate kernel on a given discrete input space, these algorithms can be used to identify ``similarities'' between two datapoints.\\\\
We are interested in studying \textit{graph kernels} $K : V \times V \rightarrow \mathbb{R}$ on the set of nodes $V$ of a graph G. In other words, for nodes $i$ and $j$ in graph G, a \textit{graph kernel} $K(i,j)$ returns a real number representing the similarity between these nodes.  
\subsection{Graph Random Features}
Despite the theoretical and empirical advantages of using kernel-based learning algorithms, the need to materialize and invert the kernel matrix leads to expensive $\mathcal{O}(N^3)$ time-complexity scaling, where $N$ is the number of nodes, or data points \cite{reid_general_2024}. To address this limitation, a Monte-Carlo mechanism called \textit{random features} was proposed to linearize kernel functions \cite{choromanski_taming_2023}. Random features use randomized functions: $\phi: \mathbb{R}^{d} \rightarrow \mathbb{R}^{s}$ to construct low-dimensional feature vectors whose dot product (a linear kernel) equals the expected value of the original kernel: 
\begin{equation}
    k(x, y) = \mathbb{E}(\phi(x)^{\top}\phi(y))
\end{equation}
These random features allow for a low-rank decomposition of the kernel matrix, which allows for better time- and space-complexity than exact kernel methods \cite{reid_quasi-monte_2023}. \\\\
However, recently a viable \textit{graph random feature} (GRF) mechanism was proposed to allow for even better kernel-based learning algorithms that enjoy a subquadratic time complexity \cite{reid_general_2024}. GRF algorithms use an ensemble of random walkers that deposit a ``load'' at every vertex they pass through that depends on: i) the product of weights of edges traversed by the walker, and ii) the marginal probability of the subwalk \cite{reid_general_2024}. With these algorithms, one can construct random features ${\phi(i)}_{i=1}^{N} \subset \mathbb{R}^{N}$ such that $\phi_{i}^{\top}\phi(j)$ gives an unbiased estimate of the $ij$-th element of the kernel matrix \cite{choromanski_taming_2023}. In other words, the algorithm estimates graph kernels by generating random features through randomized walks on the graph. \\ \\
A limitation of GRFs is that they only address a niche family of graph kernels \cite{reid_general_2024}. Thus, Choromanski et al. introduce \textit{general graph random features} (g-GRFs) which generalizes the GRF algorithm to arbitrary functions of a weighted adjacency matrix, allowing for efficient and unbiased approximations of a much broader class of graph kernels \cite{reid_general_2024}. The key contribution of g-GRF algorithms is a modulation function $f$ that controls the weight of the load deposited by random walkers as they traverse the graph based on the length of the walk. By carefully choosing and potentially learning this modulation function on a neural network, the algorithm provides an efficient, unbiased approximation of the desired graph kernel. This enables scalable kernel-based learning on graphs for a much larger class of graph kernels.

\subsection{Quasi-Monte Carlo Methods}
Quasi-Monte Carlo (QMC) sampling is a tool that improves the convergence of Monte Carlo methods. QMC methods use low-discrepancy samples to reduce integration error by replacing independent and identically distributed (IID) samples with a correlated ensemble that is deterministically constructed to be more ``diverse'' \cite{reid_repelling_2024}. By using correlated ensembles rather than IID random variables in the feature maps, one can suppress the mean squared error (MSE) of the kernel estimator \cite{reid_quasi-monte_2023}. Choromanski, Reid, and Weller propose an implementation of \textit{Quasi-Monte Carlo GRFs} (q-GRFs) by imposing \textit{antithetic termination} to correlate the lengths of \textit{random walks}, which sample a sequence of nodes connected by edges with some stopping criterion \cite{reid_quasi-monte_2023} \cite{reid_repelling_2024}. In particular, they prove that the correlations reduce the variance of estimators of the 2-regularized Laplacian kernel under mild conditions. \\ \\
In the IID implementation of GRF algorithms, each walker terminates independently with probability $p$ at every timestep \cite{reid_quasi-monte_2023}. For a pair of IID walkers for a given node $i$, this is implemented by independently sampling two \textit{termination random variables} (TRVs) between $0$ and $1$ from a uniform distribution, $t_{1, 2} \sim \mathcal{U}(0, 1)$. Each walker terminates if its respective TRV is less than $p$, i.e., $t_{1, 2} < p$. In contrast, a pair of walkers are \textit{antithetic} if their TRVs are marginally distributed as $t_{1, 2} \sim \mathcal{U}(0, 1)$ but are offset by $\frac{1}{2}$,
\begin{equation}
    t_{2} = \text{mod}_{1}\left(t_{1} + \frac{1}{2}\right),
\end{equation}
such that we have the conditional distribution
\begin{equation}
    p(t_{2}|t_{1}) = \delta\left(\text{mod}_{1}(t_{2}-t_{1}) - \frac{1}{2}\right).
\end{equation}
Since the marginal distributions over $t_{i}$ are unchanged, the estimator remains unbiased, but the
couplings between TRVs lead to statistical correlations between the walkers’ terminations. By diversifying the lengths of random walks that are sampled, preventing them from clustering together, antithetic termination thus suppresses the kernel estimator variance \cite{reid_quasi-monte_2023}. Features constructed with antithetic walkers are referred to as \textit{Quasi-Monte Carlo graph random features} (q-GRFs).

\newpage
\section{Experimental Methodology}
To experiment with the efficiency of q-GRFs, we use Python to estimate the Diffusion (or Heat), Matérn, and Inverse Cosine kernels using both the generic g-GRFs algorithm and the Quasi-Monte Carlo approach, q-GRFs. We then calculate the relative Frobenius norm $\|K - \tilde{K}\|_{F} \textfractionsolidus \|K\|_{F}$ between the true and approximate kernel matrices, where $K$ represents the true matrix and $\tilde{K}$ is the estimate. This allows us to compare the accuracy of the estimators. Below, we expand on the three kernels chosen for experimentation and their relationship to the previously studied 2-regularised Laplacian kernel. 

\subsection{Diffusion Kernels}
The \textit{Laplacian} of a graph $G$ is the negative of the matrix
\begin{equation}
    H_{ij} = \begin{cases}
            1 & \text{for } i \sim j,\\
            -d_{i} & \text{for i = j},\\
            0 & \text{otherwise,}
        \end{cases}
\end{equation}
where $d_{i}$ is the degree of vertex $i$ (i.e., the number of edges incident to vertex $i$) \cite{kondor_diffusion}. Additionally, let $D \in \mathbb{R}^{N \times N}$ be the diagonal matrix with elements $D_{ii} := \sigma_{j} W_{ij}$, the sum of edge weights connecting a vertex $i$ to its neighbors. If we denote the Laplacian of $G$ as $L$, then the \textit{normalized Laplacian} is defined as $\tilde{L} := D^{\frac{1}{2}}LD^{-\frac{1}{2}}$.\\ \\
The Diffusion (or Heat) kernel is defined analogously with physics, where equations of the form 
\begin{equation}
    \frac{d\mu}{dt} = \triangledown^{2}\mu
\end{equation}
are used to describe the Diffusion of heat and other substances through continuous media \cite{kondor_diffusion}. $\triangledown^{2}$ is defined as the \textit{Laplacian operator} on continuous spaces, where $\triangledown^{2} = \frac{\partial^2}{\partial x_{1}^2} + \frac{\partial^2}{\partial x_{2}^2} + \cdots + \frac{\partial^2}{\partial x_{N}^2}$ and $N$ is the number of vertices in the graph $G$. The Laplacian $L$ is a natural analog to the Laplacian operator, where we treat $L$ as a linear operator on vectors $\mu \in \mathbb{R}^{N}$. This motivates the \textit{discrete heat equation} on $G$,
\begin{equation}
    \frac{d\mu}{dt} = -\tilde{L}\mu
\end{equation}
This has the solution $\mu_{t} = \text{exp}(-\tilde{L}t)\mu_{0}$. Thus, the matrix 
\begin{equation}
    K_{\text{diff}}(t) := \text{exp}(-\tilde{L}t)
\end{equation}
is referred to as the \textit{Heat kernel} or \textit{Diffusion kernel}, where $t$ is the diffusion time parameter \cite{reid_quasi-monte_2023}.

\subsection{Regularized Laplacian Kernels}
If we discretize Equation $(5)$ with the backward Euler step, we have that
\begin{equation}
    \mu_{t + \delta t} = (I_{N} + \delta t \tilde{L})^{-1} \mu_{t},
\end{equation}
where $I_{N}$ is the $N \times N$ identity matrix and $\delta$ is a lengthscale parameter \cite{reid_repelling_2024}. The discrete time-evolution operator $K_{\text{lap}}^{(1)} = (I_{N} + \delta t \tilde{L})^{-1}$ is referred to as the \textit{1-regularized Laplacian kernel}, and is a member of the more general family of $d$-regularized Laplacian kernels,
\begin{equation}
    K_{\text{lap}}^{(d)} = (I_{N} + \delta t \tilde{L})^{-d}.
\end{equation}
These equations demonstrate the mathematical connection between the Diffusion (or Heat) kernel and the previously studied $d$-regularized Laplacian kernel. In their study, Choromanski, Reid, and Weller prove that q-GRFs yield lower-variance estimators of the 2-regularized Laplacian kernel under mild conditions. \cite{reid_quasi-monte_2023}. Therefore, we use previous research and this explicit mathematical connection to motivate our investigation of the Diffusion (or Heat) kernel. 

\subsection{Matérn Kernels}
The Matérn family of kernels uses a covariance function that takes as input the distances between nodes in a graph $G$. The Matérn model, i.e., the covariance function, is defined as
\begin{equation}
    M_{\nu, l}(r) = \frac{2^{1-\nu}}{\Gamma{\nu}}\left(\sqrt{2\nu}\frac{r}{l}\right)^{\nu} K_{\nu}\left(\sqrt{2\nu}\frac{r}{l}\right),
\end{equation}
where $\nu > 0$ is a smoothness parameter, $l > 0$ is a length scale parameter, $r$ is the distances between a pair of nodes, $\Gamma$ is the gamma function, and $K_{\nu}$ is a modified Bessel function of the second kind of order $\nu$ \cite{porcu_matern_2023}.\\ \\
When $\nu = k + \frac{1}{2}$, with $k$ being a nonnegative integer, the Matérn covariance function simplifies into the product of an exponential and a polynomial of degree $k$ \cite{porcu_matern_2023}. In general, for $k \in \mathbb{N}^{+}$,
\begin{equation}
    M_{k + \frac{1}{2}, l} = \text{exp}\left(-\frac{\sqrt{2k + 1}r}{l}\right)\frac{k!}{(2k)!}\sum_{i=0}^{k} \frac{(k+i)!}{i!(k-i)!}\left(\frac{2\sqrt{2k+1}r}{l}\right)^{k-i}
\end{equation}
We choose to explore the performances of q-GRFs using a kernel defined by the simplified Matérn covariance function for $\nu = 2.5$ (corresponding to $k = 2$) and $l = 1$.

\subsection{Inverse Cosine Kernels}
The matrix for the inverse cosine kernel is defined as 
\begin{equation}
    K_{\text{inv. cos}} = \text{cos}\left(\frac{\tilde{L}\pi}{4}\right)
\end{equation}
The inverse cosine kernel treats lower complexity functions almost equally, with a significant reduction in the upper end of the spectrum \cite{goos_kernels_2003}.

\subsection{Implementation}
All of our empirical work is performed in Python, and the code can be found in the \href{https://github.com/QMCSoftware/QMCSoftware}{QMCSoftware} GitHub repository. Our code is based off of previous work done by Choromanski, Reid, and Weller, whose Python code is published in the \href{https://github.com/isaac-reid/antithetic_termination}{Antithetic Termination} GitHub repository \cite{reid_quasi-monte_2023}. 
\\ \\In our preliminary experimentation, we use g-GRFs and q-GRFs to generate unbiased estimates of the Gram matrix that represents each corresponding kernel function (namely, the Diffusion, Matérn, and Inverse Cosine kernels). We then compute the relative Frobenius norm $\|K - \tilde{K}\|_{F} \textfractionsolidus \|K\|_{F}$ between the true and approximate kernel matrices to compare the quality of our estimators. For the q-GRFs algorithm, we employ ensembles of antithetic walkers defined by Equations $(2)$ and $(3)$ to marginally distribute the termination random variables. This approach prevents clustering by diversifying the lengths of the random walks. Initially, for each graph that we perform our first set of tests on, we consider $2, 4, 8, $ and $16$ walks, taking $100$ repeats for the variance of the approximation error, and we use the termination probability $p = 0.5$. Section 4 describes our preliminary analysis on different graph kernels. After analyzing our results, we pivot and focus on studying characteristics of graphs types that benefit from antithetic termination. Section 6 summarizes our second set of experiments.
\newpage
\section{Preliminary Results}
Figures 1, 2, 3, and 4 present our preliminary results with four different kernels (Diffusion with $t = 0.5$, 2-Regularized Laplacian, Matérn with $\nu = 2.5$ and $l = 1$, and Inverse Cosine). We perform our initial experiment on a broad class of graphs: small Erdos-Rényi, large Erdos-Rényi, Binary tree, Ladder, and four real-world examples (``Karate,'' ``Dolphins,'' ``Football,'' and ``Eurosis'') from Ivashkin's \href{https://github.com/vlivashkin/community-graphs}{community graphs} GitHub repository \cite{ivashkin2016logarithmic}. Choromanski, Reid, and Weller's research (investigating the 2-regularized Laplacian kernel) involved experimentation on these same eight graphs. We aim to find an alternative kernel function that achieves the same improved variance estimation when applying antithetic termination techniques.  
\begin{figure}[htp]
                \centering
                \begin{large}
                \textbf{Figure 1: Diffusion (Heat) Kernel}
                \end{large}
            \includegraphics[scale = 0.35]{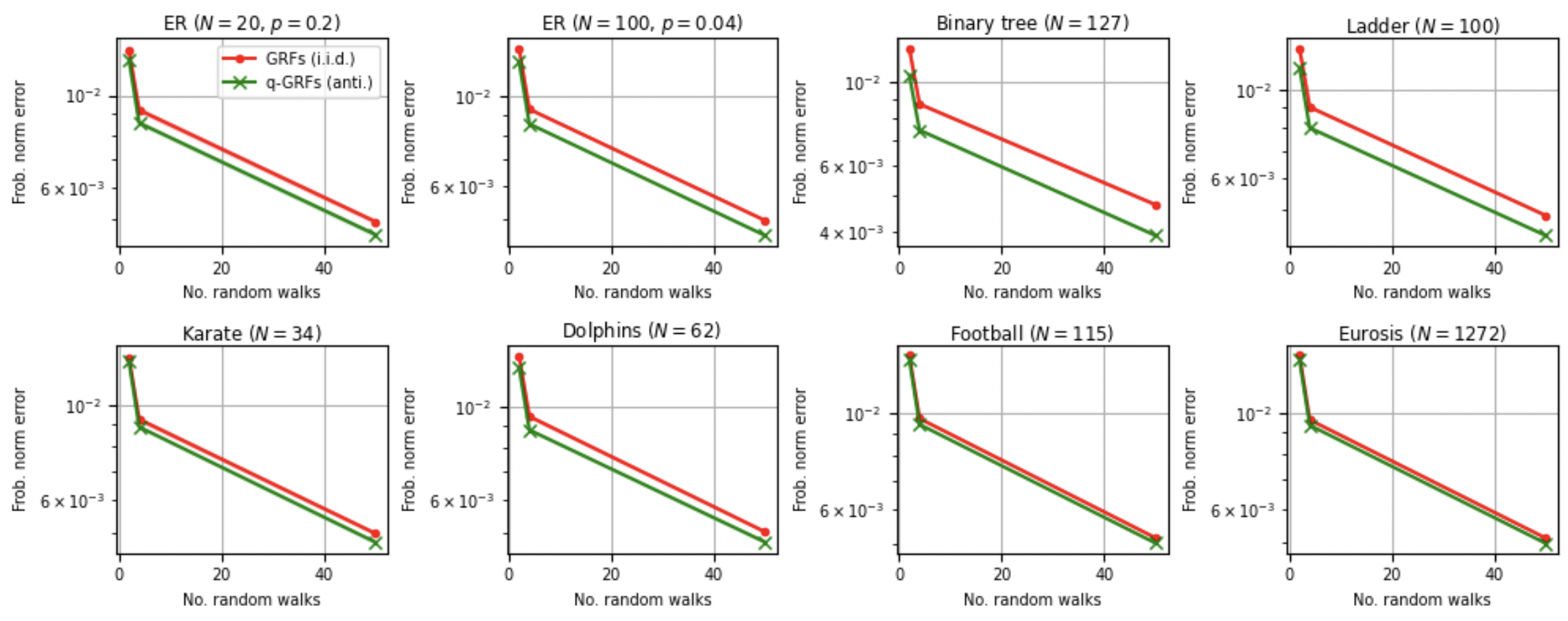}
                \caption{Relative Frobenius norm error of estimators of the Diffusion (or heat) kernel with $t = 0.5$ using general GRFs (red circles) and q-GRFs (green crosses). Lower is better. $N$ gives the number of nodes and $p$ is the edge-generation probability for the Erdös-Rényi graphs. One standard deviation is shaded, but in some of the graphs it is too small to easily see. The novel q-GRFs algorithm performed better on all eight graphs, mirroring the behavior of the previosuly studied 2-regularized Laplacian kernel \cite{reid_quasi-monte_2023}. For the sake of demonstrating long-term behavior, we visualize 50 random walks where clearly q-QRFs yield a lower variance estimator for the Diffusion kernel as number of random walks increases. 
                }
                \label{fig: diffusion_1}
            \end{figure}
\begin{figure}[htp]
                \centering
                \begin{large}
                \textbf{Figure 2: Regularized Laplacian Kernel}
                \end{large}
                \includegraphics[scale = 0.43]{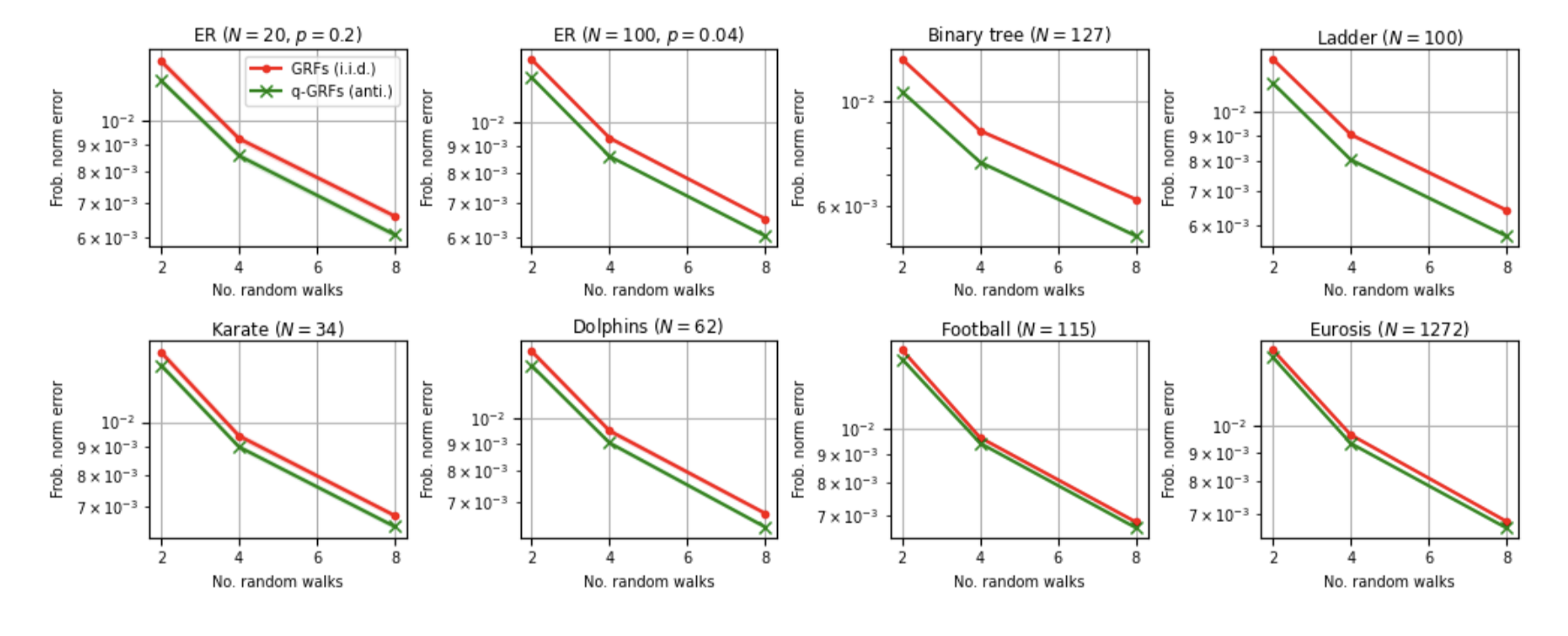}
                \caption{Relative Frobenius norm error of estimators of the 2-regularized Laplacian kernel using GRFs (red circles) and q-GRFs (green crosses). Lower is better. $N$ gives the number of nodes and $p$ is the edge-generation probability for the Erdös-Rényi graphs. One standard deviation is shaded, but it is too small to easily see. These graphs are a direct result of a previous study by Choromanski, Reid, and Weller that introduced Quasi-Monte Carlo antithetic termination on random walks. These results serve as a baseline for comparing alternative kernel functions.}
                \label{fig: laplacian_1}
            \end{figure}
\begin{figure}[htp]
                \centering
                \begin{large}
                \textbf{Figure 3: Matérn Kernel}
                \end{large}
                \includegraphics[scale = 0.6]{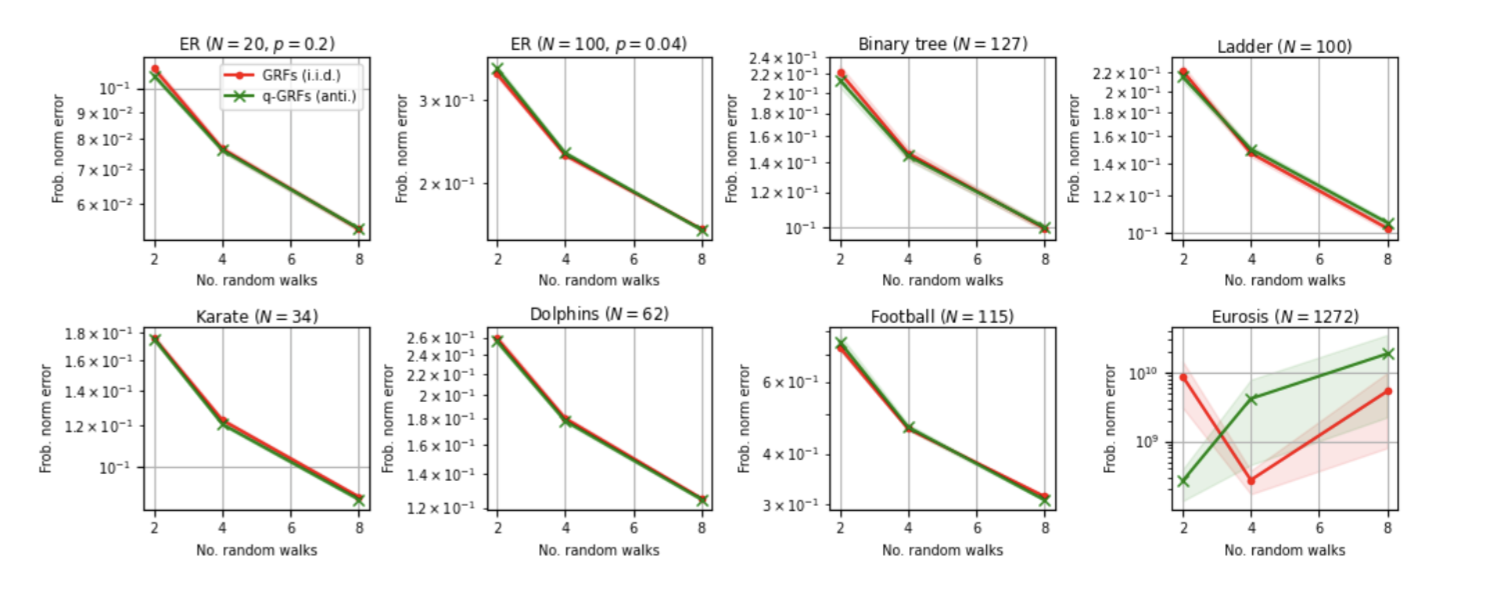}
                \caption{Relative Frobenius norm error of estimators of the Matérn kernel with smoothness parameter $\nu = 2.5$ and length scale parameter $l = 1$ using GRFs (red circles) and q-GRFs (green crosses). Lower is better. $N$ gives the number of nodes and $p$ is the edge-generation probability for the Erdös-Rényi graphs. One standard deviation is shaded, but in some of the graphs it is too small to easily see. Q-GRFs performed worse or, at best, the same as regular g-GRFs. }
                \label{fig: matern_1}
            \end{figure}
\begin{figure}[htp]
                \centering
                \begin{large}
                \textbf{Figure 4: Inverse Cosine Kernel}
                \end{large}
                \includegraphics[scale = 0.6]{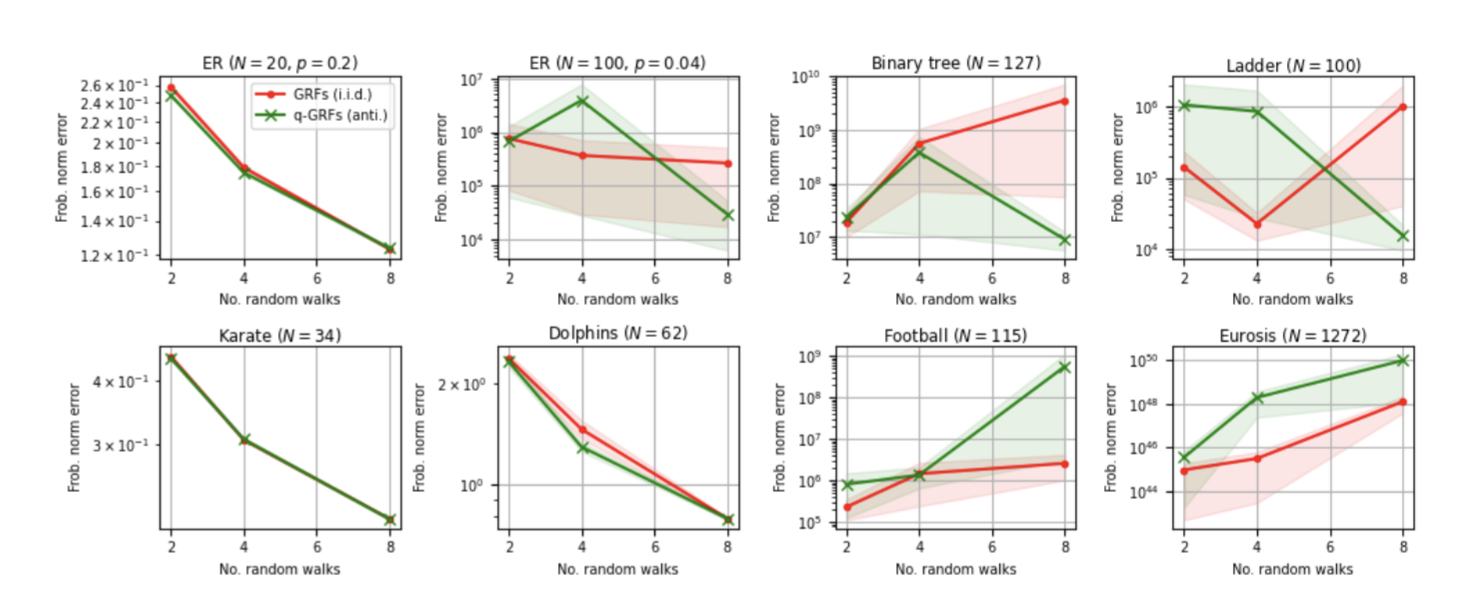}
                \caption{Relative Frobenius norm error of estimators of the inverse cosine kernel using GRFs (red circles) and q-GRFs (green crosses). Lower is better. $N$ gives the number of nodes and $p$ is the edge-generation probability for the Erdös-Rényi graphs. One standard deviation is shaded, but in some of the graphs it is too small to easily see. The q-GRFs algorithm performed best on Erdös-Rényi, Binary Tree, Ladder, and Dolphins graphs.}
                \label{fig: inv_cos_1}
            \end{figure}

\subsection{Preliminary Conclusions}
In most cases, the quality of the kernel approximations naturally improves with the number of walkers. However, q-QRFs failed to consistently yield lower variance estimators for the matérn and inverse cosine kernels on all eight graphs. The Diffusion kernel, however, was the only other arbitrary function that mirrored the results previously established by Choromanski, Reid, and Weller. Therefore, due to the inherent randomness of the walk-based algorithm, we repeated our experimentation by running seven test cases to better quantify the success rate of q-GRFs on estimating Diffusion (or Heat) kernels.\\\\
For the Erdös-Rényi graphs where $N$=20, q-GRFs either performed similarly to g-GRFs (IID) or yielded lower variance estimators in all seven test cases. However, when $N$ was increased to 100, only six out of seven test cases yielded lower variance estimators for the Diffusion kernel. For Binary tree graphs, q-GRFs performed better than g-GRFs (IID) in three test cases. Moreover, antithetic termination achieved lower variance estimators in all seven test cases for Ladder graphs. Finally, the four real-world graphs Karate, Dolphins, Football, and Eurosis had varying results, yet all consistently generated lower variance estimators in most test cases. In one out of seven test cases, q-GRFs achieved lower variance estimators of the Diffusion kernel on all eight graphs simultaneously (Figure 1).\\\\
This result prompts us to pivot our research and investigate what characterizes graphs that benefit from antithetic termination; can q-QRFs produce lower variance estimators of the Diffusion kernel for a particular graph type? 

\section{Graph Experimentation}
Due to the effects of randomness on our preliminary results, we narrow our work to identify specific graph types that successfully yield lower variance estimators of Diffusion kernels. This section outlines the theoretical background of the graph families with which we perform the bulk of our experimentation following our initial analysis. We examine Erdős-Rényi and Barabási-Albert random graph models, Binary Trees, and Ladder graphs. In the subsequent section, we provide visualizations and analysis of our experimentation with the Diffusion kernel.

\subsection{Erdős-Rényi Graphs}
In the Erdős-Rényi model, a graph is constructed by connecting labeled nodes randomly \cite{bollobas_evolution_1984}. Each edge is included in the graph with probability $p$, independent of the other edges. Thus, the probability for generating each graph that has $n$ nodes and $M$ edges is
\begin{equation}
    \text{prob}(p, n, M) = p^{M}(1-p)^{{n \choose 2 - M}}
\end{equation}
The parameter $p$ acts as a weighting function: as $p$ increases from $0$ to $1$, the model becomes more likely to include graphs with more edges and less likely to include graphs with fewer edges \cite{bollobas_evolution_1984}.\\ \\
We thus continue our experimentation by comparing the performance of q-GRFs with g-GRFs on different Erdős-Rényi models. We test multiple Erdős-Rényi graphs by changing the \textit{spin} parameter, i.e., the number of vertices. Specifically, we test Erdős-Rényi graphs with spins of $20$, $60$, and $100$. Our Erdős-Rényi graphs are generated based on code from the \href{https://github.com/tomdbar/eco-dqn/tree/master?tab=readme-ov-file}{GitHub repository} created by Barrett, Clements, Foerster, and Lvovsky \cite{barrett_tomdbareco-dqn_2024}.

\subsection{Barabási-Albert Graphs}
The Barabási–Albert (BA) model is an algorithm for generating random scale-free networks, i.e., complex graphs that contain non-trivial topological features that often occur in networks representing real systems. These complex networks describe a wide variety of systems in nature and society such as the cell, a network of chemicals linked by chemical reactions, and the Internet, a network of routers and computers connected by physical links \cite{albert_statistical_2002}. The BA model incorporates the concepts of \textit{growth}, meaning the number of nodes in the network increases over time; and \textit{preferential attachment}, meaning that the more connected a node is, the more likely it is to receive new links.\\ \\
The BA algorithm uses a parameter $m \in \mathbb{N}^{+}$. The network initializes with a network of $m_{0} \geq m$. Then, at each step, it adds one new node and then samples $m$ existing vertices from the network, with a probability that is proportional to the number of links that the existing nodes already have \cite{albert_statistical_2002}. The formula for the probability $p_{i}$ that the new node is connected to node $i$ is
\begin{equation}
    p_{i} = \frac{k_{i}}{\sum_{j} k_{j}},
\end{equation}
where $k_{i}$ is the degree of node $i$ and the sum is made over all pre-existing nodes $j$ \cite{albert_statistical_2002}.\\ \\
We thus study the performance of q-GRFs and g-GRFs on Barabási-Albert graphs with spin parameters (i.e., number of vertices) of $20$, $60$, and $100.$ Our BA graphs are generated based on code from the \href{https://github.com/tomdbar/eco-dqn/tree/master?tab=readme-ov-file}{GitHub repository} created by Barrett, Clements, Foerster, and Lvovsky \cite{barrett_tomdbareco-dqn_2024}.

\subsection{Binary Trees}
To define binary trees, we must first define some other terminology commonly used in graph theory:
\begin{itemize}
    \item A \textit{rooted tree} is a tree with a designated vertex called the \textit{root}, and each edge is implicitly directed away from the root.
    \item In a rooted tree, if vertex $v$ immediately precedes vertex $w$ on the path from the root to $w$, then $v$ is a \textit{parent} of $w$ and $w$ is a \textit{child} of $v$.
    \item An \textit{ordered tree} is a rooted tree in which the children of each vertex are assigned a fixed ordering.
    \item An $n$-ary tree ($n \geq 2$) is a rooted tree in which every vertex has $n$ or fewer children.
\end{itemize}
Now we can use these terms to define binary trees as a special type of $2$-ary tree. Specifically, a binary tree is an ordered $2$-ary tree in which each child is designated either a \textit{left-child} or a \textit{right-child} \cite{slides_binary_tree}. The \textit{left} (or \textit{right}) \textit{subtree} of a vertex $v$ in a binary tree is the binary subtree spanning the left (or right)-child of $v$ and all of its descendants. Thus, the designation of left-child or right-child means that two different binary trees may be indistinguishable when regarded more generally as ordered trees \cite{slides_binary_tree}. Binary trees can be ``balanced'' or ``unbalanced'' depending on the heights of the left and right subtrees for each node. If the height of the left subtree is greater than that of the right subtree, then the graph is \textit{left-heavy}; if the height of the left subtree is less than that of the right subtree, then the graph is \textit{right-heavy}; and if the heights of both subtrees are equal, then the graph is balanced \cite{karlton_performance_1976}.\\ \\
In Python, we use a built-in "binarytree" library to generate balanced Binary Tree graphs. We test the Diffusion kernel on binary trees with $2$, $4$, $8$, $10$, $20$, $50$, and $100$ random walks.

\subsection{Ladder Graphs}
We next move on to studying kernel approximations of ladder graphs, again using the Diffusion (or heat) kernel. A ladder graph $L_{n}$ is defined to be a planar, undirected graph with $2n$ vertices and $3n-2$ edges, and can be obtained as the Cartesian product of two path graphs, one of which has only one edge: $L_{n, 1} = P_{n} \times P_{2}$ \cite{weisstein_ladder_nodate}. A ladder graph $L_{n}$ looks like a ladder with $n$ rungs. We perform experiments on ladder graphs with $8$, $9$, $10$, and $11$ rungs and simulate $2$, $4$, $8$, $10$, $20$, $50$, and $100$ random walks. In Python, we use the library "NetworkX" to generate the ladder graphs with varying numbers of rungs.

\newpage
\section{Results with Diffusion Kernels}
\subsection{Erdős-Rényi Experimentation}
These graphs are generated based on code from Barrett's GitHub repository \cite{barrett_tomdbareco-dqn_2024}. In our experiments, we test various \textit{spin} parameters to see if the number of vertices bears any pragmatic significance on improving q-GRFs performance on Erdős-Rényi graphs. 
\begin{figure}[htp]
                \centering
                \begin{large}
                \textbf{Figure 5: Erdős-Rényi Graphs, Spin=20}
                \end{large}
                \includegraphics[scale = 0.35]{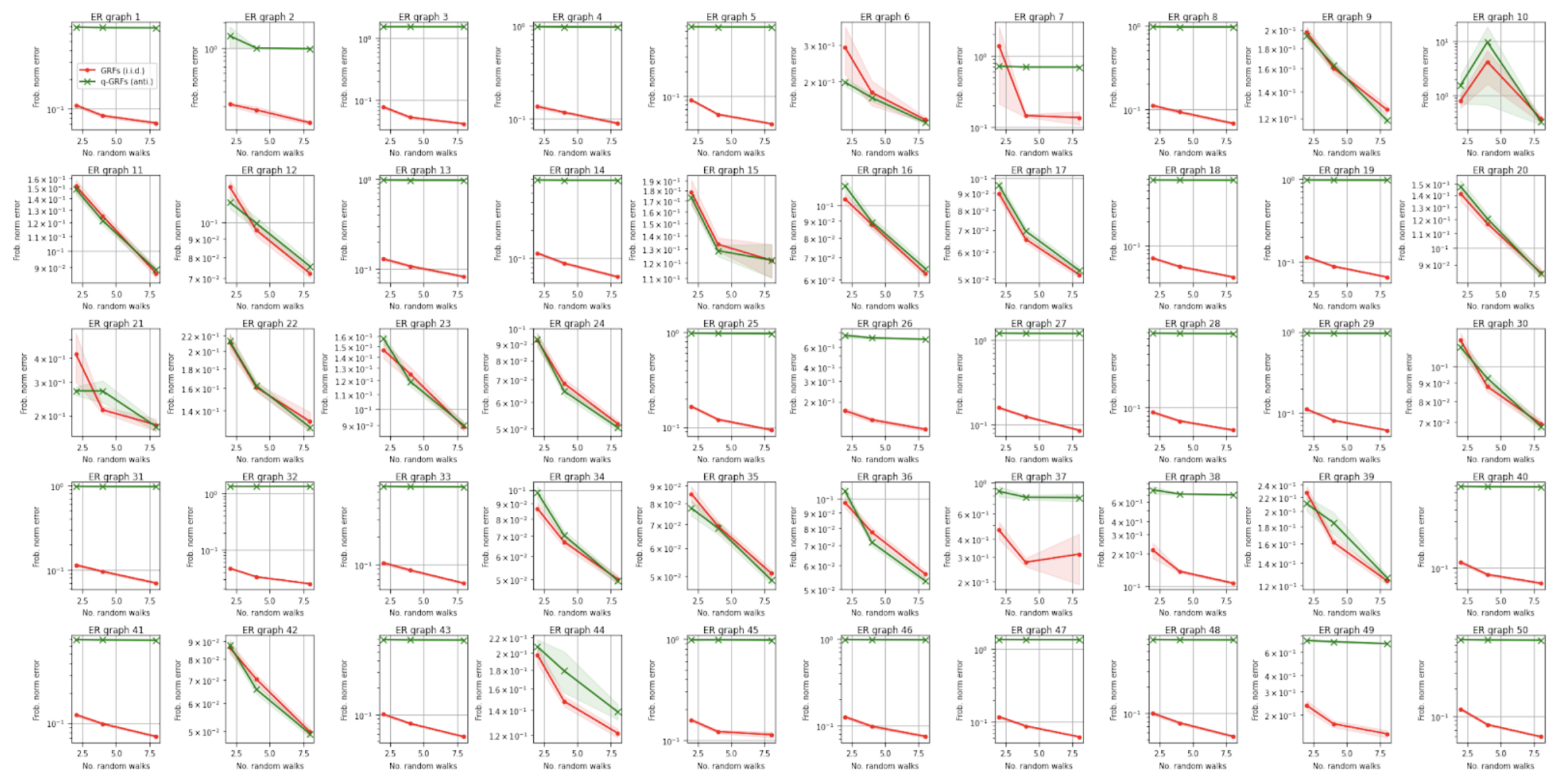}
                \caption{50 Erdős-Rényi graphs with a spin parameter set to 20. Q-GRF's yield lower variance estimators of the Diffusion kernel 26\%\ of the time. }
                \label{fig: ER_20}
            \end{figure}
\begin{figure}[htp]
                \centering
                \begin{large}
                \textbf{Figure 6: Erdős-Rényi Graphs, Spin=60}
                \end{large}
            \includegraphics[scale = 0.4]{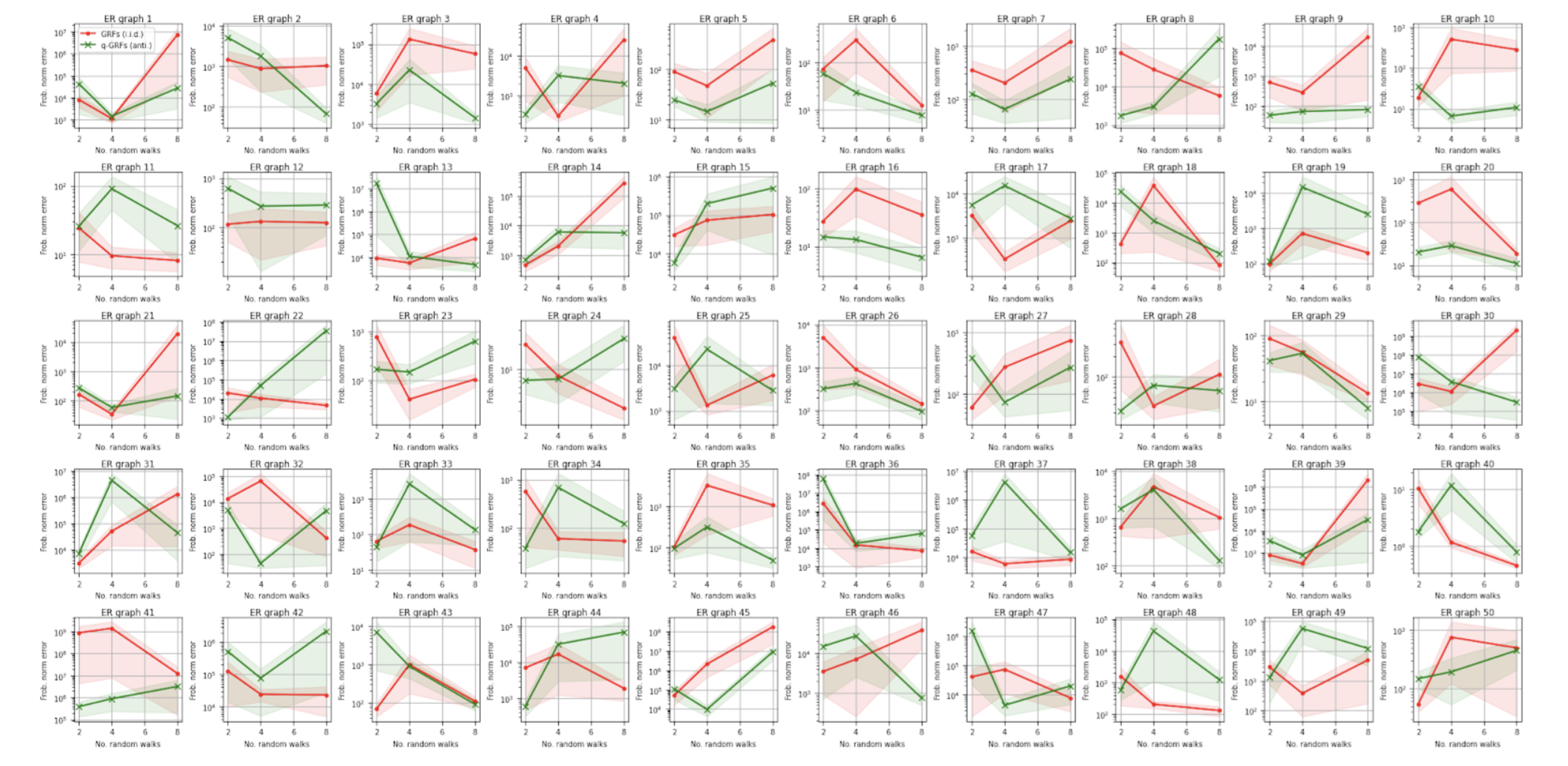}
                \caption{50 Erdős-Rényi graphs with a spin parameter set to 60. Q-GRF's yield lower variance estimators of the Diffusion kernel 58\%\ of the time.}
                \label{fig: ER_60}
            \end{figure}
\begin{figure}[htp]
                \centering
                \begin{large}
                \textbf{Figure 7: Erdős-Rényi Graphs, Spin=100}
                \end{large}
            \includegraphics[scale = 0.36]{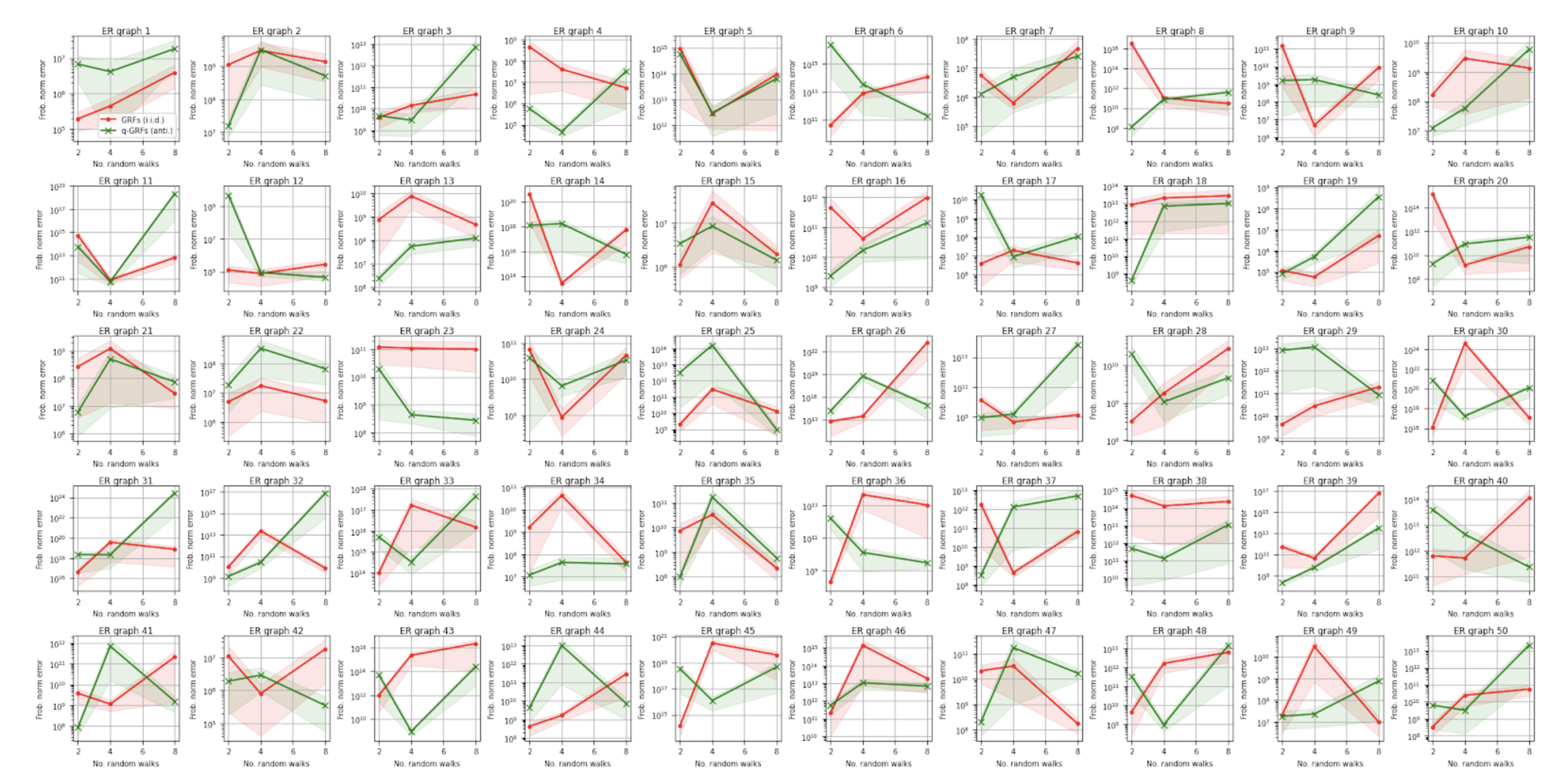}
                \caption{50 Erdős-Rényi graphs with a spin parameter set to 100. Q-GRF's yield lower variance estimators of the Diffusion kernel 54\%\ of the time.}
                \label{fig: ER_100}
            \end{figure}
\subsection{Barabási-Albert Experimentation}
These graphs are generated based on code from Barrett's GitHub repository \cite{barrett_tomdbareco-dqn_2024}. In our experiments, we test various \textit{spin} parameters to see if the number of vertices bears any pragmatic significance on improving q-GRFs performance on Barabási-Albert graphs.
\begin{figure}[htp]
                \centering
                \begin{large}
                \textbf{Figure 8: Barabási-Albert Graphs, Spin=20}
                \end{large}
        \includegraphics[scale = 0.33]{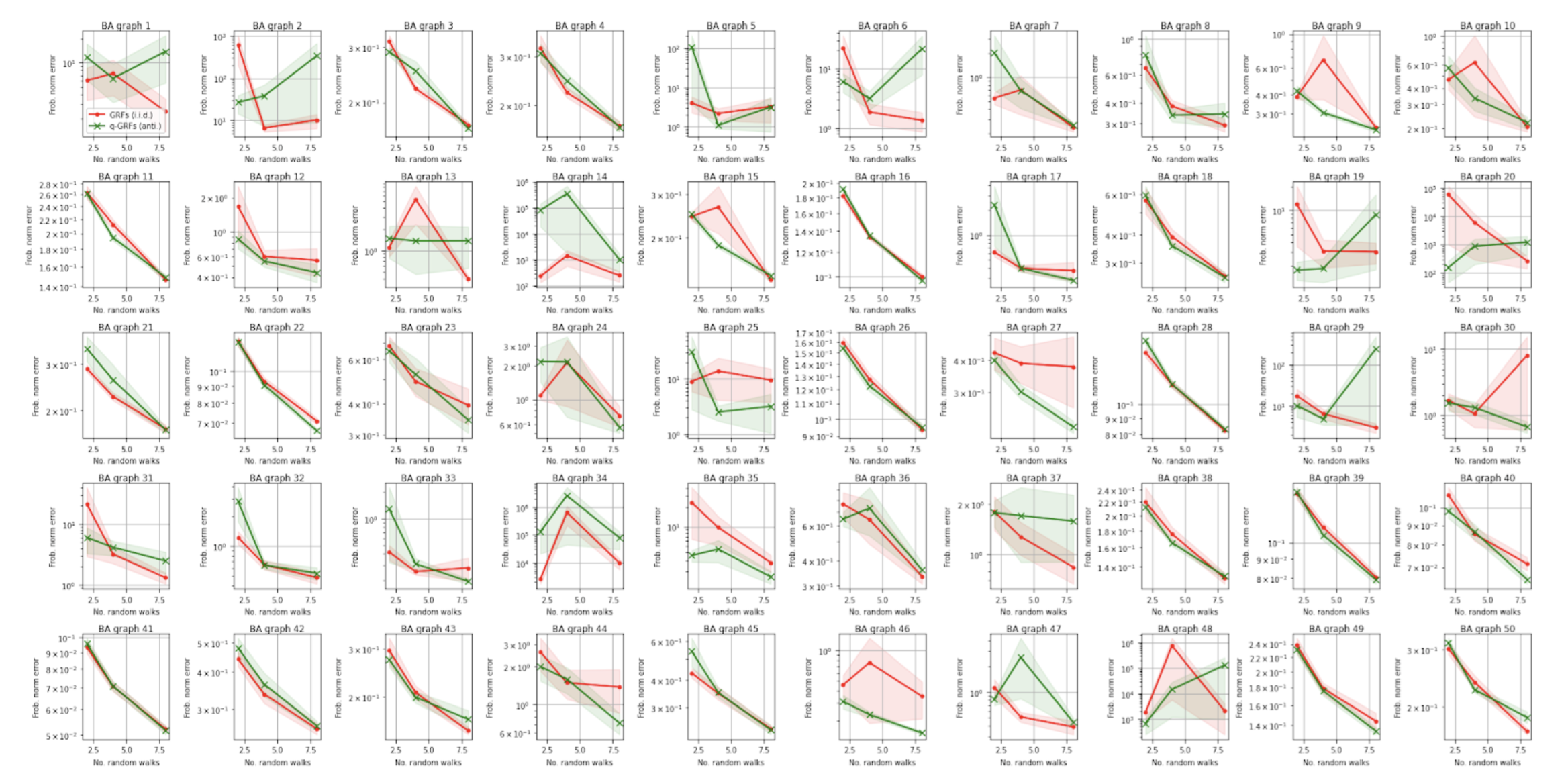}
                \caption{50 Barabási-Albert graphs with a spin parameter set to 20. Q-GRF's yield lower variance estimators of the Diffusion kernel 44\%\ of the time. }
                \label{fig: BA_20}
            \end{figure}
\begin{figure}[htp]
                \centering
                \begin{large}
                \textbf{Figure 9: Barabási-Albert Graphs, Spin=60}
                \end{large}
        \includegraphics[scale = 0.4]{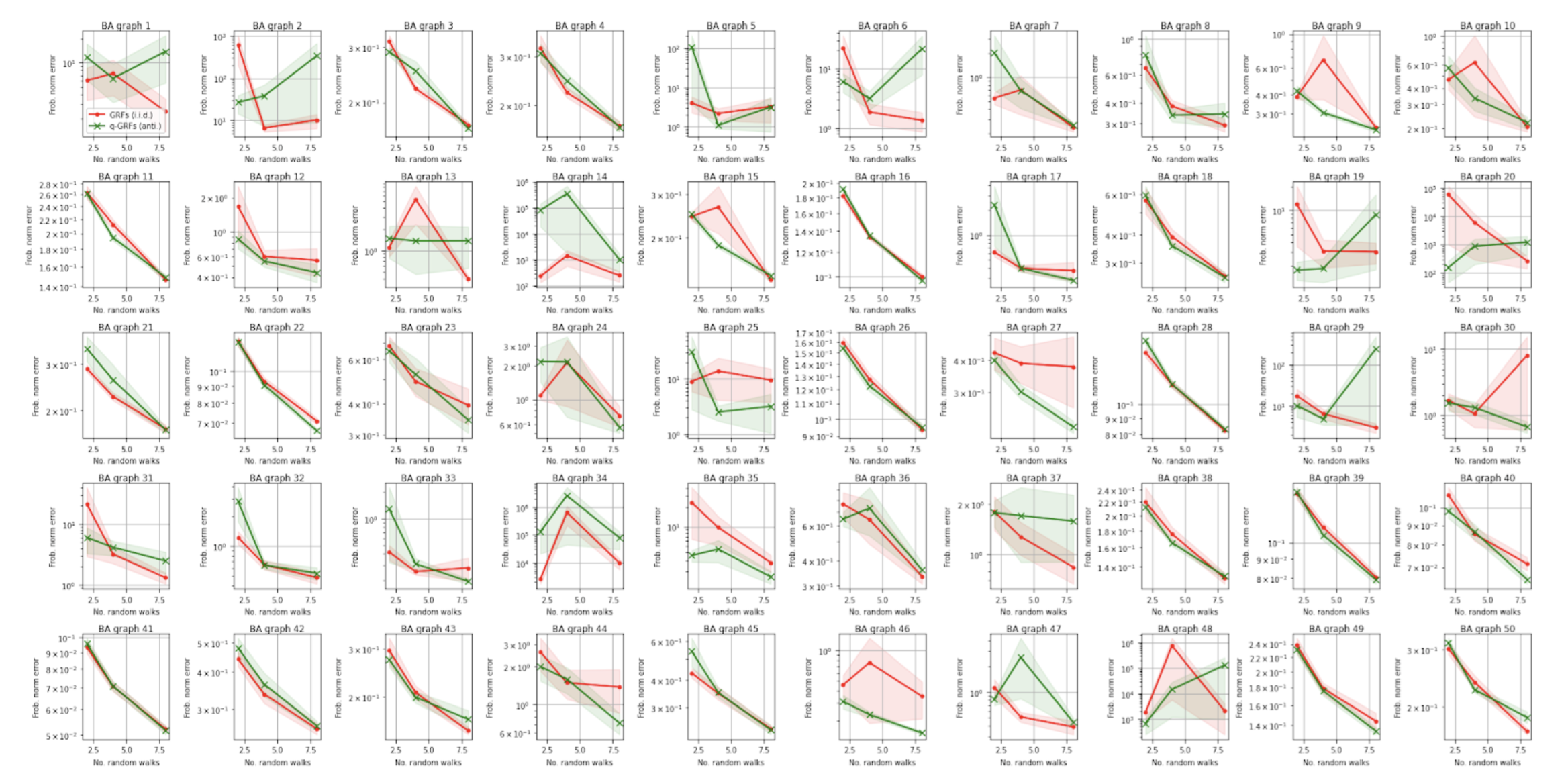}
                \caption{50 Barabási-Albert graphs with a spin parameter set to 60. Q-GRF's yield lower variance estimators of the Diffusion kernel 46\%\ of the time. }
                \label{fig: BA_60}
            \end{figure}
\begin{figure}[htp]
                \centering
                \begin{large}
                \textbf{Figure 10: Barabási-Albert Graphs, Spin=100}
                \end{large}
        \includegraphics[scale = 0.4]{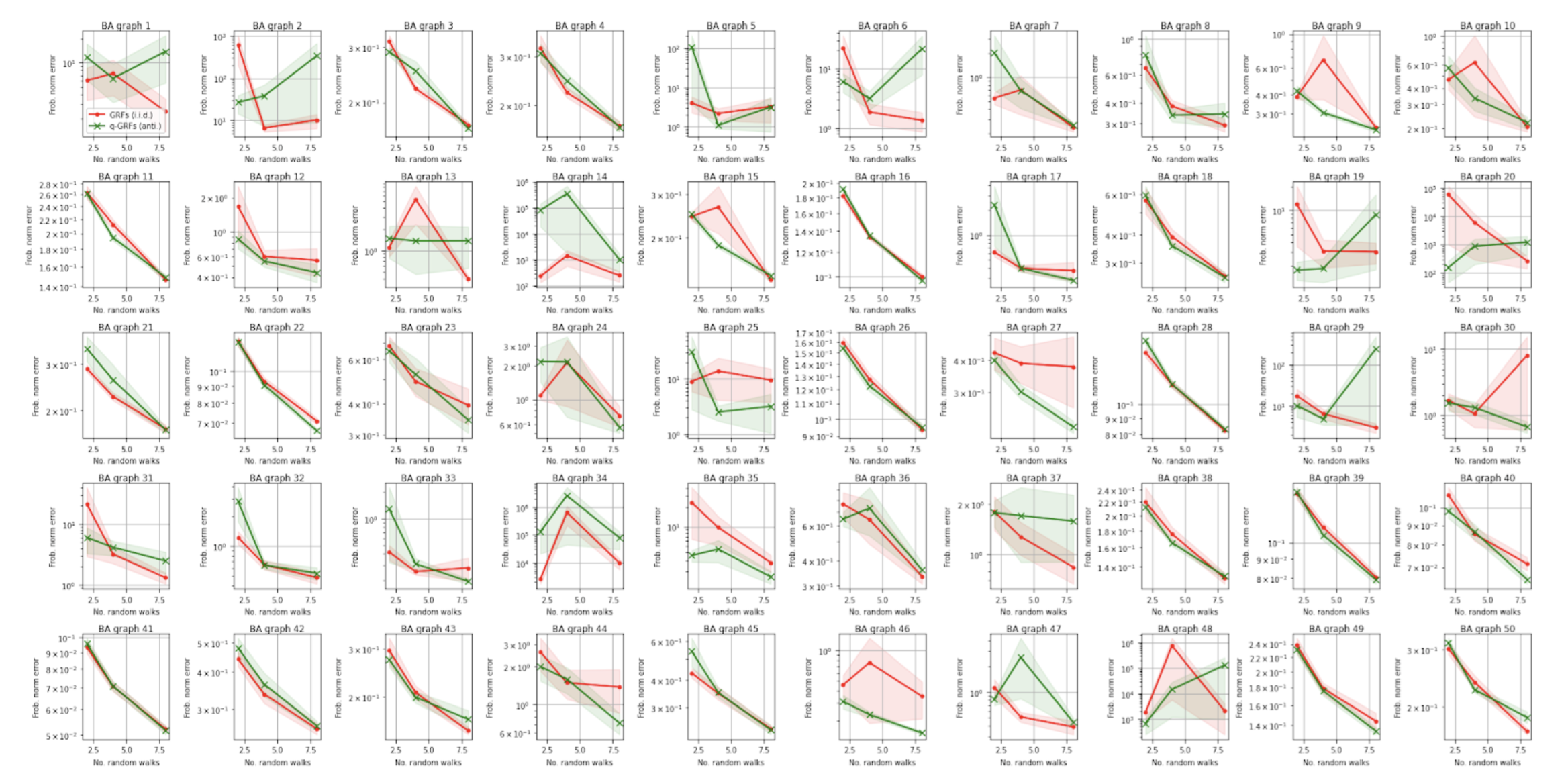}
                \caption{50 Barabási-Albert graphs with a spin parameter set to 100. Q-GRF's yield lower variance estimators of the Diffusion kernel 44\%\ of the time. }
                \label{fig: BA_100}
            \end{figure}
\newpage
\subsection{Binary Tree Experimentation}
These graphs are generated in Python using the "binarytree" library which allows one to generate, visualize, inspect and manipulate binary trees.
\begin{figure}[htp]
                \centering
                \begin{large}
                \textbf{Figure 11: Binary Tree Graph 1}
                \end{large}
                \includegraphics[scale = 0.52]{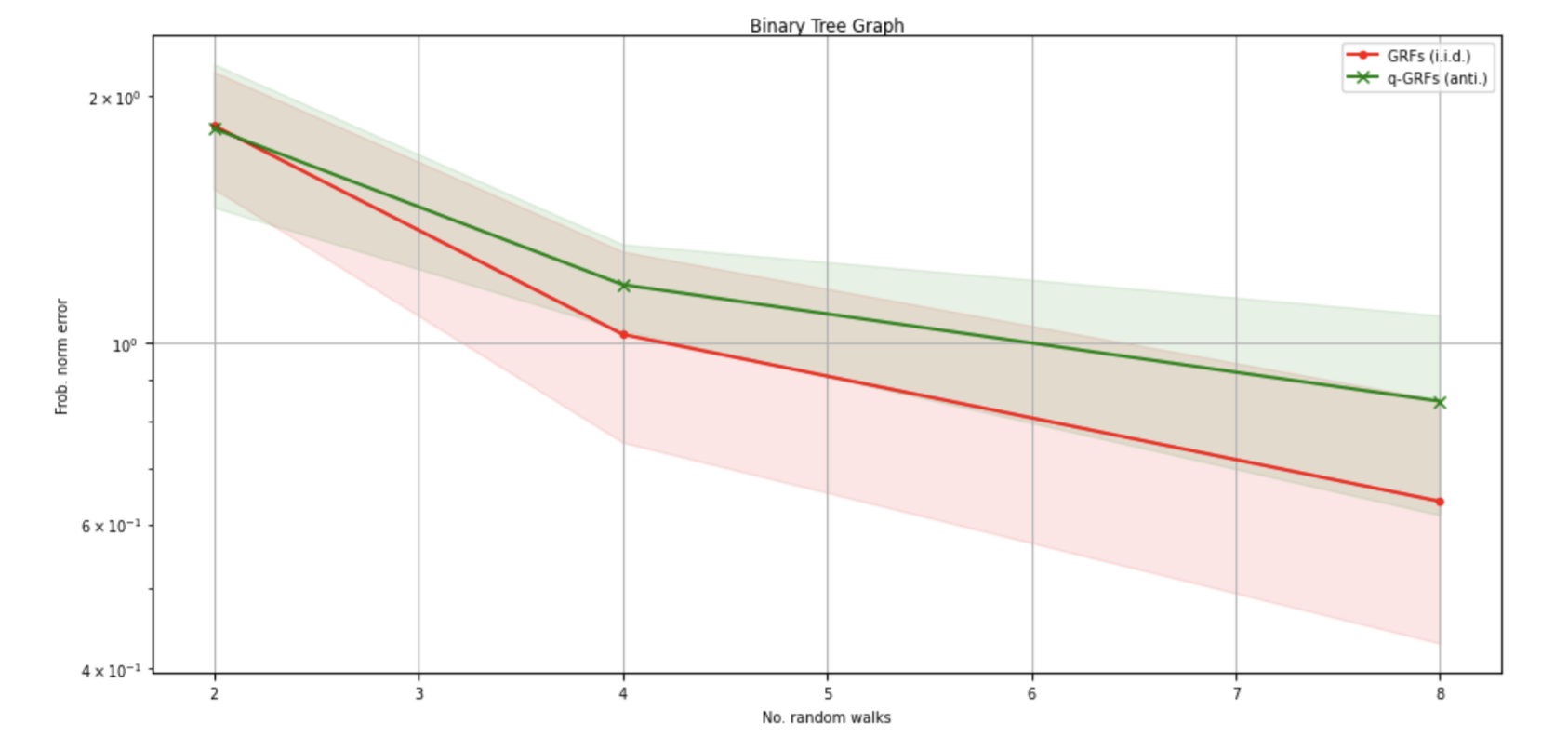}
                \caption{Binary Tree graph with number of random walks set to (2, 4, 8). Clearly, q-GRFs fail to yield lower variance estimators of the Diffusion kernel.}
                \label{fig: Binary_Tree}
            \end{figure}
\begin{figure}[htp]
                \centering
                \begin{large}
                \textbf{Figure 12: Binary Tree Graph 2}
                \end{large}
                \includegraphics[scale = 0.52]{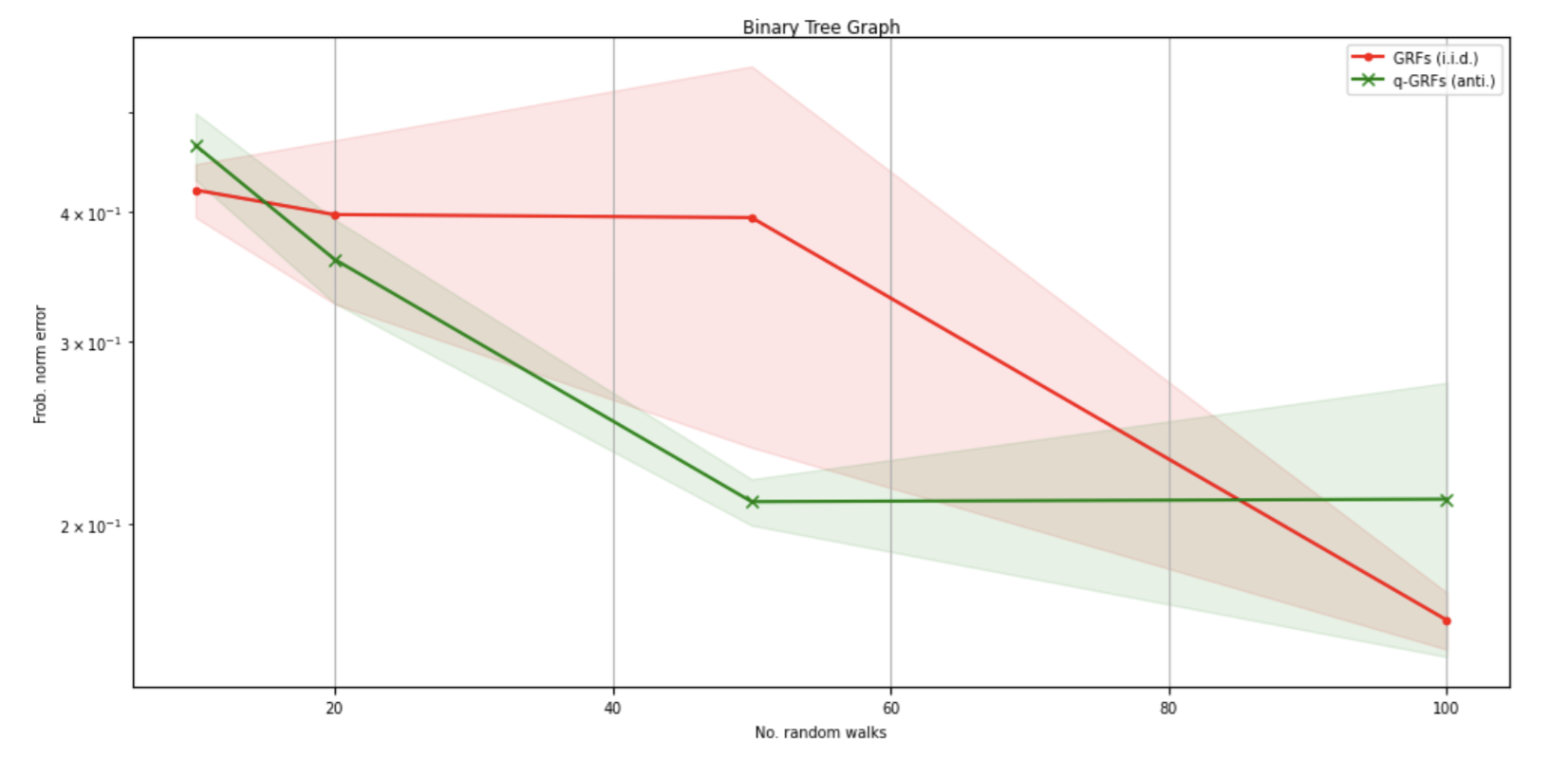}
                \caption{Binary Tree graph with number of random walks set to (10, 20, 50, 100) Clearly, q-GRFs fail to yield lower variance estimators of the Diffusion kernel even as number of walks increases.}
                \label{fig: Binary_Tree}
            \end{figure}
\newpage
\subsection{Ladder Graph Experimentation}
These graphs are generated in Python using the "NetworkX" library which allows one to create, manipulate, and study the structure, dynamics, and functions of complex networks. We test Ladder graphs with $8$, $9$, $10$, and $11$ rungs, but only those with $9$ and $10$ rungs show an improvement in lower variance estimators of the Diffusion kernel using q-GRFs.
\begin{figure}[htp]
                \centering
                \begin{large}
                \textbf{Figure 13: Ladder Graph 1 (9 Rungs)}
                \end{large}
                \includegraphics[scale = 0.46]{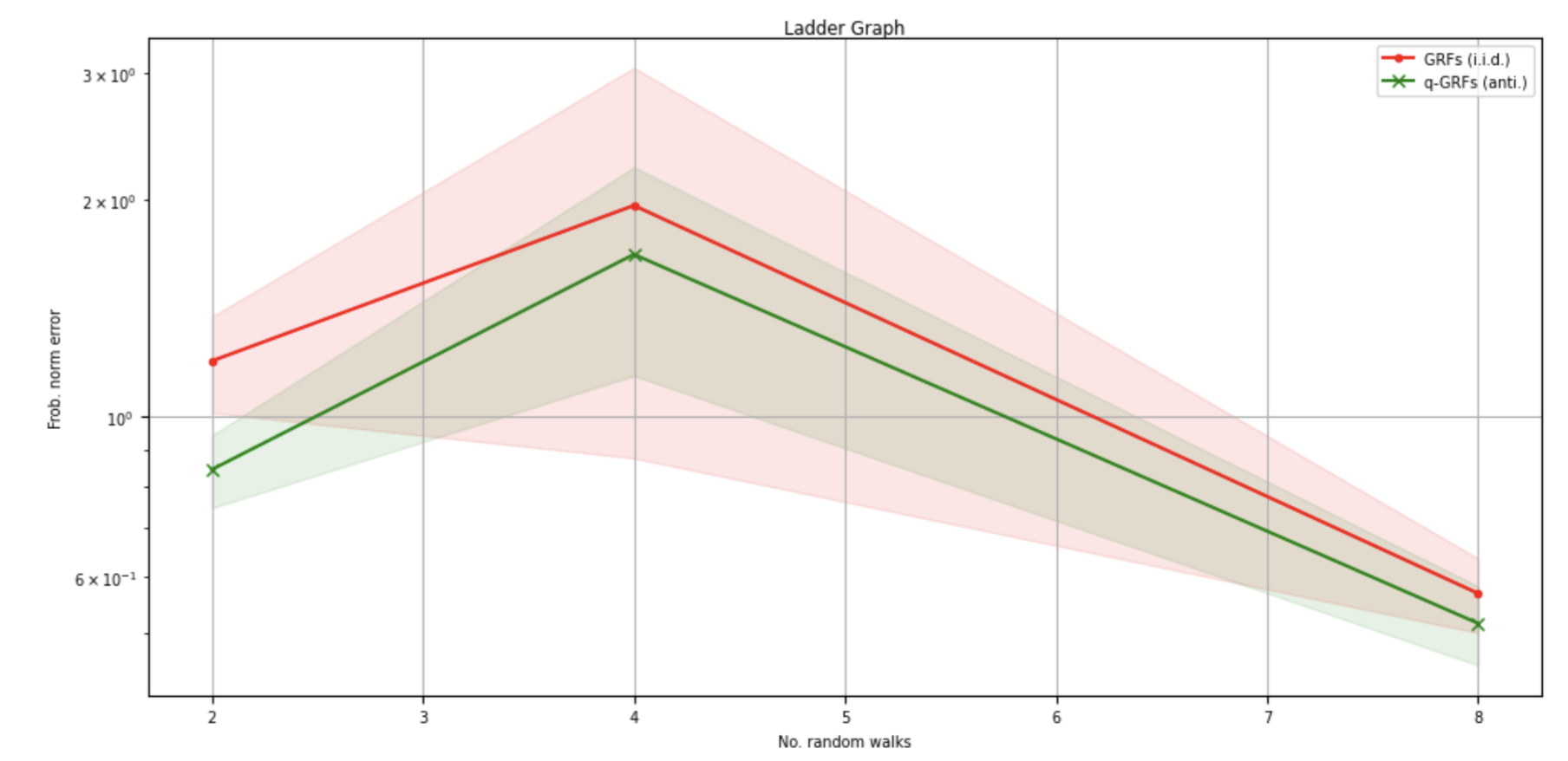}
                \caption{Ladder graph with number of random walks set to (2, 4, 8) and number of rungs set to 9. Q-GRFs yield lower variance estimators of the Diffusion kernel.}
                \label{fig: Ladder1_graph}
            \end{figure}
\begin{figure}[htp]
                \centering
                \begin{large}
                \textbf{Figure 14: Ladder Graph 2 (9 Rungs)}
                \end{large}
                \includegraphics[scale = 0.46]{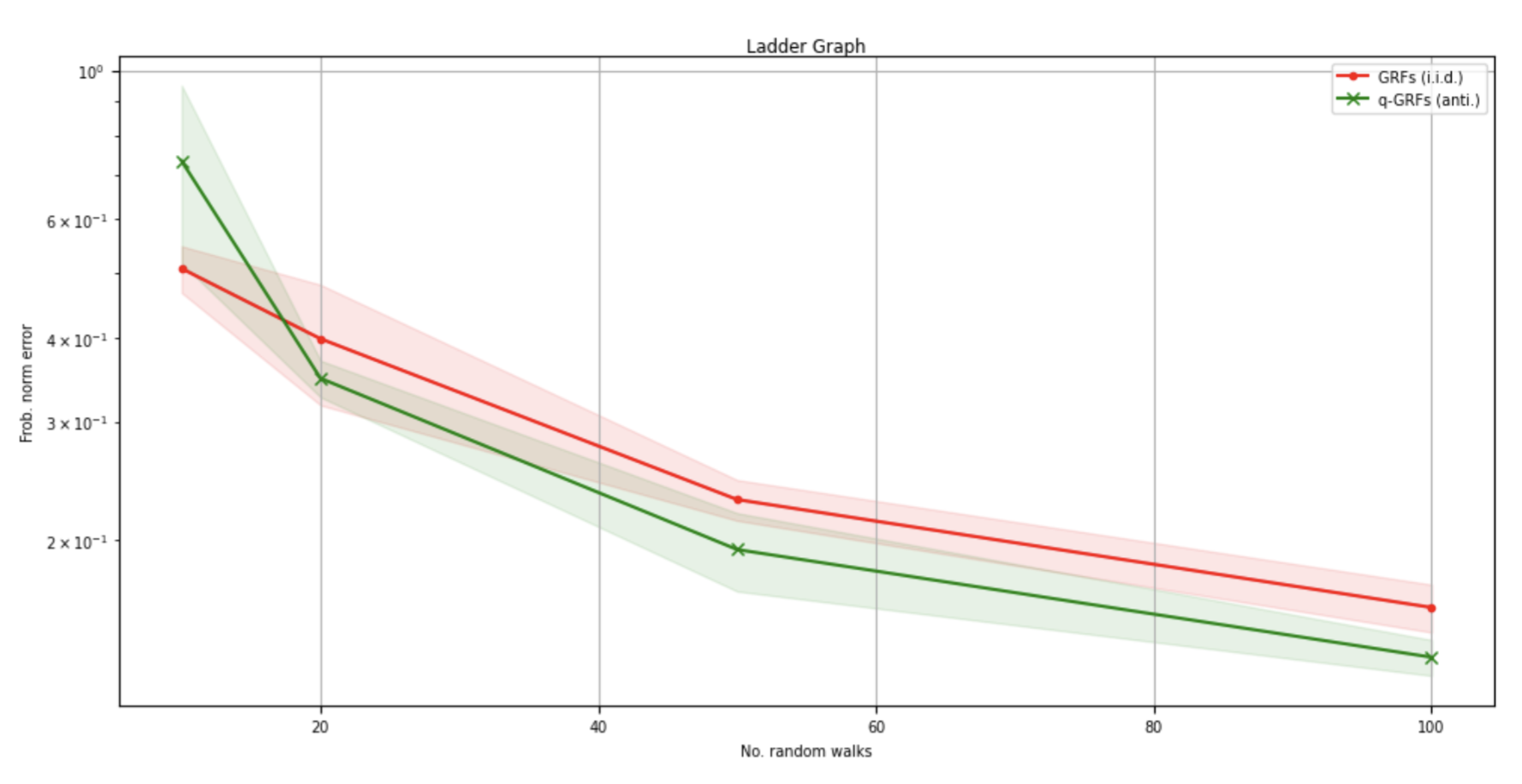}
                \caption{Ladder graph with number of random walks set to (10, 20, 50, 100) and number of rungs set to 9. Q-GRFs yield lower variance estimators of the Diffusion kernel.}
                \label{fig: Ladder2_graph}
            \end{figure}
\begin{figure}[htp]
                \centering
                \begin{large}
                \textbf{Figure 15: Ladder Graph 3 (10 Rungs)}
                \end{large}
                \includegraphics[scale = 0.46]{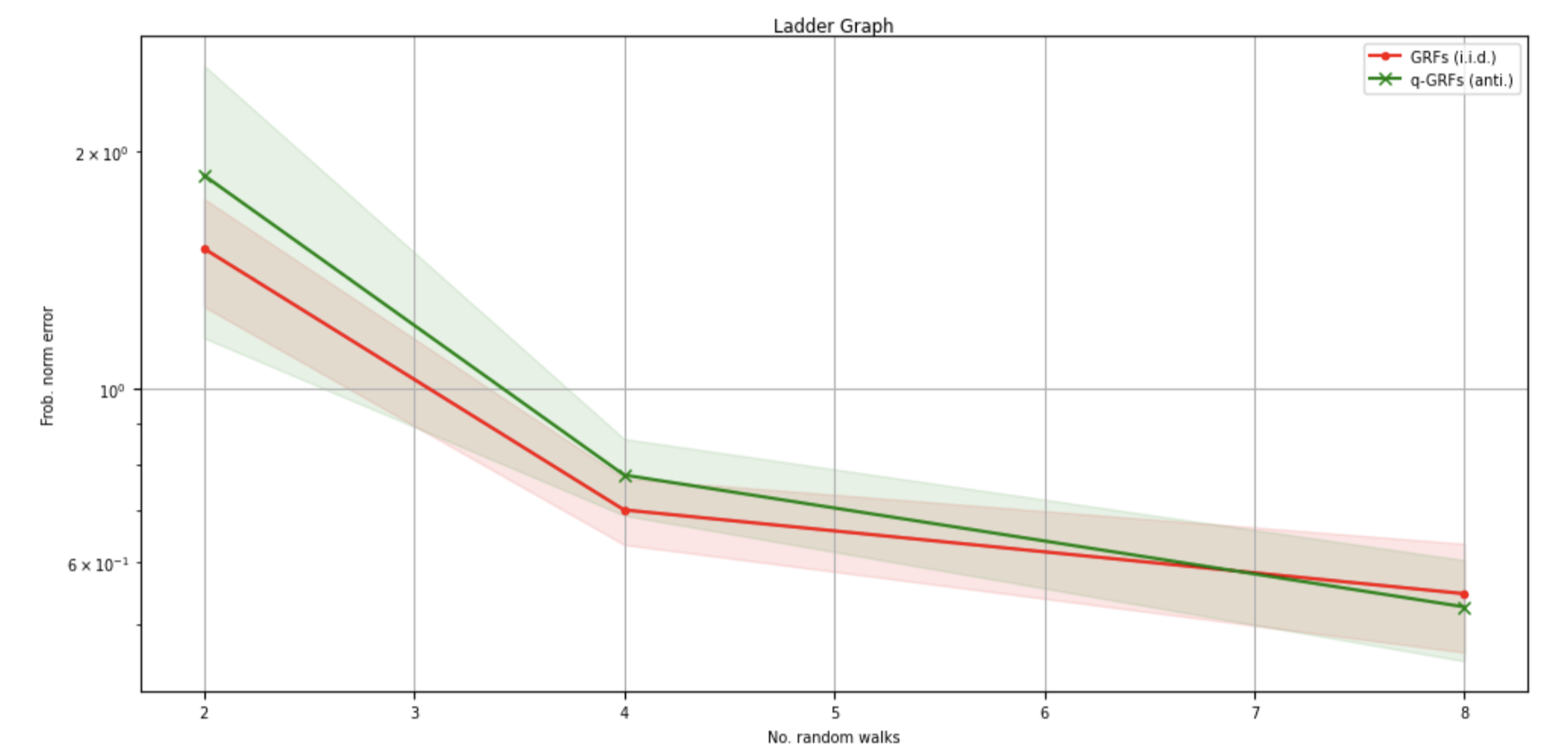}
                \caption{Ladder graph with number of random walks set to (2, 4, 8) and number of rungs set to 10. Q-GRFs yield lower variance estimators of the Diffusion kernel.}
                \label{fig: Ladder3_graph}
            \end{figure}
\begin{figure}[htp]
                \centering
                \begin{large}
                \textbf{Figure 16: Ladder Graph 4 (10 Rungs)}
                \end{large}
                \includegraphics[scale = 0.46]{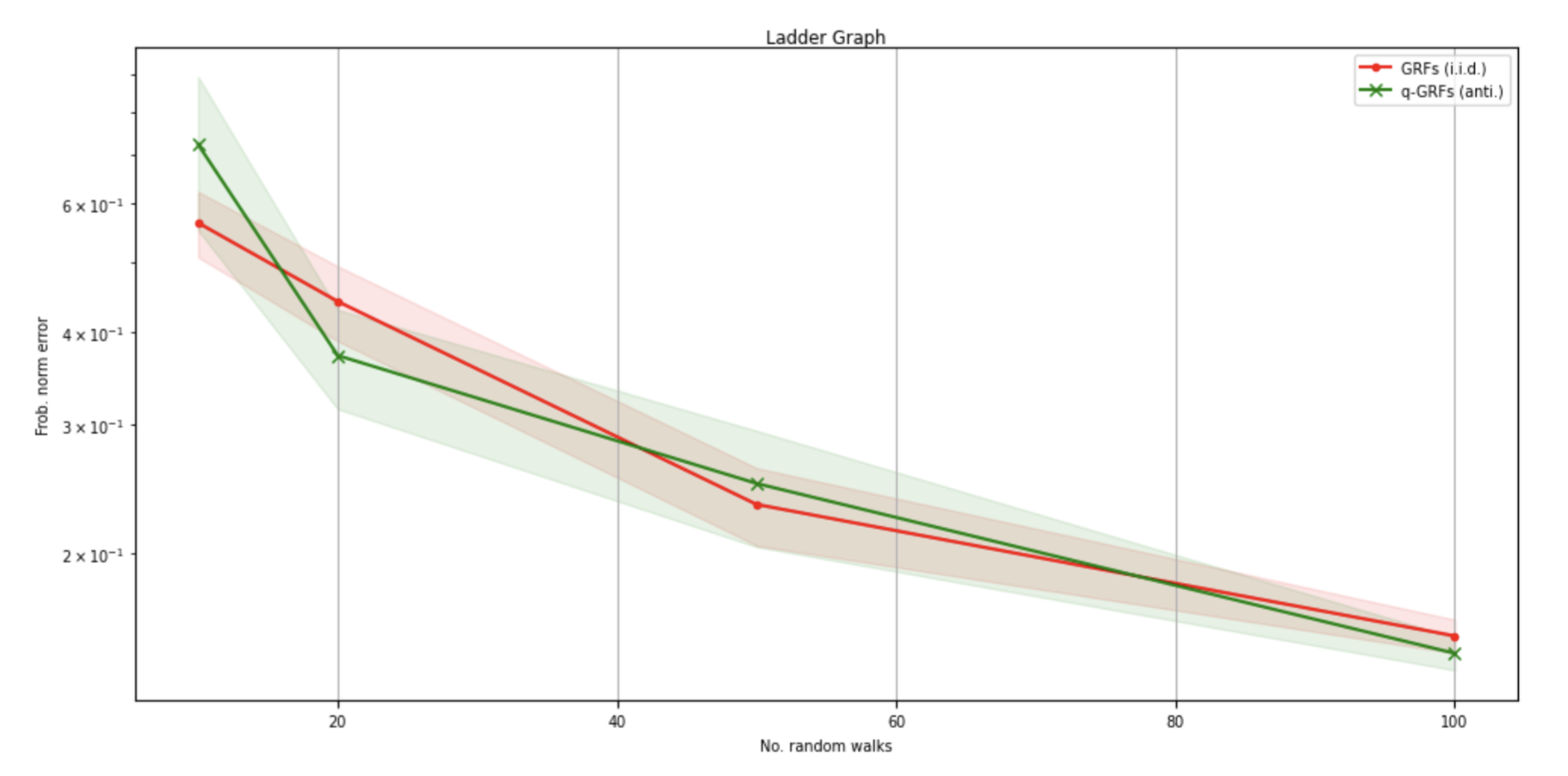}
                \caption{Ladder graph with number of random walks set to (10, 20, 50, 100) and number of rungs set to 10. Q-GRFs yield lower variance estimators of the Diffusion kernel.}
                \label{fig: Ladder4_graph}
            \end{figure}
\newpage
\section{Final Thoughts}
\subsection{Conclusion}
With both the Erdős-Rényi and Barabási-Albert graph types, increasing the spin parameter did not improve the overall accuracy of q-GRFs in achieving lower variance estimators of Diffusion kernels. In fact, the middle rate, with spin = 60, had the highest success in both random graph models, with the ER graphs showing a 58\%\ success rate (with 29 out of the 50 graphs showing improvement) and the BA graphs showing a 46\%\ success rate (with 23 out of the 50 graphs showing an improvement). Neither of these rates is high enough to conclude at this time that the spin parameter has any effect on antithetic termination when applied to random graph models.\\\\ 
For the Binary Tree graphs, q-GRFs failed to yield lower variance estimators of the Diffusion kernel, regardless of the number of walkers. We conclude that for a balanced binary tree, q-GRFs (antithetic termination) does not yield lower varience estimators of the Diffusion kernel when compared to the standard g-GRFs (IID) algorithm.\\\\
For Ladder graphs with $9$ and $10$ rungs, we assert that q-QRFs do yield lower variance estimators of the Diffusion kernel. Pictorially, the green line representing antithetic termination converges to a smaller bound than the red line for IID in both (2, 4, 8) and (10, 20, 50, 100) random walks (Figures 13-16). However, the number of rungs on the Ladder graph significantly impacts the algorithm's performance. When the number of rungs is set to $8$ or $11$, antithetic termination fails to outperform g-GRFs (not pictured). Therefore, we claim that the number of rungs affects whether quasi graph random features (q-GRFs) achieve lower variance estimators of Diffusion kernel on ladder graphs-–theoretical results explaining and investigating this phenomenon are forthcoming.
\subsection{Future Work}
We investigate various kernel families and graph types in extension to Choromanski, Reid, and Weller’s previously proposed class of quasi-Monte Carlo graph random features (q-GRFs) for the unbiased estimation of graph kernel matrices. This work builds upon some of the earliest quasi-Monte Carlo methods for kernels defined on combinatorial objects, paving the way for future exploration into kernels defined on discrete structures.\\\\
Our work indicates a promising direction for future research. In our preliminary experiments, we tested several kernels beyond the 2-regularized Laplacian to evaluate if q-GRFs achieve lower variance estimators for other arbitrary kernel functions. While in some cases the Diffusion kernel yields comparable results to the previously proposed 2-regularized Laplacian kernel, the inherent randomness in the q-GRFs algorithm leads to inconsistent outcomes. We cannot confidently assert that q-QRFs consistently yield lower variance estimators of the Diffusion kernel. However, it is evident that the algorithm has the potential to achieve lower-error estimates, though future theoretical work investigating its inherent randomized nature is necessary to better understand the behavior of antithetic termination on Diffusion kernels.\\\\
Additionally, we have begun investigating the characteristics of graphs that particularly benefit from antithetic termination. While our work has primarily focused on the Diffusion kernel–-with the aim of replicating previous results obtained using the 2-regularized Laplacian kernel–-it is imperative to test alternative graph types. This will help determine if any other combinations of graph types and kernel functions achieve a similar lower variance estimator to the 2-regularized Laplacian kernel when applying q-GRFs. In the context of our presented work, further theoretical research is necessary to fully understand how the number of rungs on a ladder graph affects the performance of antithetic termination. Moreover, error analysis quantifying and visualizing the performance of q-GRFs on the Diffusion and 2-regularized Laplacian kernel with various graph types is essential for improving the efficiency of this kernel-based learning algorithm. 

\newpage
\section{Relative Contributions and Acknowledgements}
Our work builds upon the foundational research conducted by Choromanski, Reid, and Weller \cite{reid_quasi-monte_2023}. We acknowledge their pivotal contributions in devising the antithetic termination procedure and expanding quasi-Monte Carlo methods to combinatorial objects. \\\\ 
Furthermore, we thank Dr. Fred Hickernell and Dr. Yuhan Ding for their feedback and ongoing support throughout the SURE: Summer Undergraduate Research Experience program hosted at the Illinois Institute of Technology in Chicago. We also acknowledge the support and funding provided by NSF-DMS-2244553.















\pagebreak
\newpage
\printbibliography

@book{goos_kernels_2003,
	location = {Berlin, Heidelberg},
	title = {Kernels and Regularization on Graphs},
	volume = {2777},
	isbn = {978-3-540-40720-1 978-3-540-45167-9},
	pages = {144--158},
	booktitle = {Learning Theory and Kernel Machines},
	publisher = {Springer Berlin Heidelberg},
	author = {Smola, Alexander J. and Kondor, Risi},
	editor = {Schölkopf, Bernhard and Warmuth, Manfred K.},
	editorb = {Goos, Gerhard and Hartmanis, Juris and Van Leeuwen, Jan},
	editorbtype = {redactor},
	date = {2003},
	langid = {english},
	doi = {10.1007/978-3-540-45167-9_12},
	note = {Series Title: Lecture Notes in Computer Science}
}

@inproceedings{kondor_diffusion,
  title={Diffusion kernels on graphs and other discrete structures},
  author={Risi Kondor},
  booktitle={International Conference on Machine Learning},
  year={2002},
  url={https://api.semanticscholar.org/CorpusID:8606662}
}

@article{reid_quasi-monte_2023,
	title = {Quasi-Monte Carlo Graph Random Features},
	doi = {10.48550/arXiv.2305.12470},
	number = {{arXiv}:2305.12470},
	publisher = {{arXiv}},
	author = {Reid, Isaac and Choromanski, Krzysztof and Weller, Adrian},
	date = {2023-05-21}
}

@article{reid_general_2024,
	title = {General Graph Random Features},
	url = {http://arxiv.org/abs/2310.04859},
	doi = {10.48550/arXiv.2310.04859},
	number = {{arXiv}:2310.04859},
	publisher = {{arXiv}},
	author = {Reid, Isaac and Choromanski, Krzysztof and Berger, Eli and Weller, Adrian},
	urldate = {2024-07-19},
	date = {2024-05-24},
	eprinttype = {arxiv},
	eprint = {2310.04859 [cs, stat]}
}

@article{reid_repelling_2024,
	title = {Repelling Random Walks},
	doi = {10.48550/arXiv.2310.04854},
	number = {{arXiv}:2310.04854},
	publisher = {{arXiv}},
	author = {Reid, Isaac and Berger, Eli and Choromanski, Krzysztof and Weller, Adrian},
	date = {2024-05-24}
}

@article{choromanski_taming_2023,
	title = {Taming graph kernels with random features},
	doi = {10.48550/arXiv.2305.00156},
	number = {{arXiv}:2305.00156},
	publisher = {{arXiv}},
	author = {Choromanski, Krzysztof},
	date = {2023-04-28}
}

@misc{porcu_matern_2023,
	title = {The Matérn Model: A Journey through Statistics, Numerical Analysis and Machine Learning},
        doi = {10.48550/arXiv.2303.02759},
        number = {{arXiv}:2303.02759},
	publisher = {{arXiv}},
	shorttitle = {The Mat{\textbackslash}'ern Model},
	author = {Porcu, Emilio and Bevilacqua, Moreno and Schaback, Robert and Oates, Chris J.},
	date = {2023-03-05},
	langid = {english}
}

@inproceedings{ivashkin2016logarithmic,
  title={Do logarithmic proximity measures outperform plain ones in graph clustering?},
  author={Ivashkin, Vladimir and Chebotarev, Pavel},
  booktitle={International Conference on Network Analysis},
  pages={87--105},
  year={2016},
  organization={Springer}
}

@article{barrett_tomdbareco-dqn_2024,
	title = {Exploratory Combinatorial Optimization with Reinforcement Learning},
	doi = {10.48550/arXiv.1909.04063},
	number = {{arXiv}:1909.04063},
	publisher = {{arXiv}},
	author = {Barrett, Thomas D. and Clements, William R. and Foerster, Jakob N. and Lvovsky, A. I.},
	date = {2020-01-31}
}

@article{bollobas_evolution_1984,
	title = {The evolution of random graphs},
	volume = {286},
	issn = {0002-9947, 1088-6850},
	doi = {10.1090/S0002-9947-1984-0756039-5},
        pages = {257--274},
	number = {1},
	journaltitle = {Transactions of the American Mathematical Society},
	shortjournal = {Trans. Amer. Math. Soc.},
	author = {Bollobás, Béla},
	date = {1984},
	langid = {english}
}

@online{weisstein_ladder_nodate,
	title = {Ladder Graph},
	rights = {Copyright 1999-2024 Wolfram Research, Inc.  See https://mathworld.wolfram.com/about/terms.html for a full terms of use statement.},
	url = {https://mathworld.wolfram.com/},
	type = {Text},
	author = {Weisstein, Eric W.},
	urldate = {2024-07-24},
	langid = {english},
	note = {Publisher: Wolfram Research, Inc.}
}

@misc{slides_binary_tree,
    title = {Graph Theory – Lecture 4: Trees},
    howpublished = {\url{https://www.cs.columbia.edu/~cs4203/files/GT-Lec4.pdf}},
    Institution = {Columbia University in the City of New York},
    note = {Accessed: 2024-07-25}
}

@article{karlton_performance_1976,
	title = {Performance of height-balanced trees},
	volume = {19},
	issn = {0001-0782, 1557-7317},
	doi = {10.1145/359970.359989},
        pages = {23--28},
	number = {1},
	journaltitle = {Communications of the {ACM}},
	shortjournal = {Commun. {ACM}},
	author = {Karlton, P. L. and Fuller, S. H. and Scroggs, R. E. and Kaehler, E. B.},
	date = {1976-01},
	langid = {english}
}

@article{albert_statistical_2002,
	title = {Statistical mechanics of complex networks},
	volume = {74},
	issn = {0034-6861, 1539-0756},
	doi = {10.1103/RevModPhys.74.47},
        pages = {47--97},
	number = {1},
	journaltitle = {Reviews of Modern Physics},
	shortjournal = {Rev. Mod. Phys.},
	author = {Albert, Reka and Barabasi, Albert-Laszlo},
	date = {2002-01-30}
}

\end{document}